\documentclass[10pt,twocolumn,letterpaper]{article}

\usepackage{iccv}
\usepackage{times}
\usepackage{epsfig}
\usepackage{graphicx}
\usepackage{amsmath}
\usepackage{amssymb}

\usepackage{subfigure}
\usepackage{algorithm}
\usepackage{algorithmic}
\usepackage{placeins}

\usepackage[pagebackref=true,breaklinks=true,letterpaper=true,colorlinks,bookmarks=false]{hyperref}

\iccvfinalcopy 

\ificcvfinal\fi
\begin{document}

\title{Deep Learning for Object Saliency Detection and Image Segmentation}

\author{\it Hengyue Pan$^\dagger$\qquad  Bo Wang$^\ddagger$\thanks{This work was done at York University, Canada.}\qquad Hui Jiang$^\dagger$ \\
$^\dagger$ Department of Electrical Engineering and Computer Science\\
Lassonde School of Engineering, York University, 4700 Keele Street, Toronto, Canada\\
$^\ddagger$ Department of Computer Science, Stanford University, California, USA  \\
{\tt panhy@cse.yorku.ca, wangbo.yunze@gmail.com, hj@cse.yorku.ca}
}

\maketitle

\begin{abstract}
In this paper, we propose several novel deep learning methods for object saliency detection based on the powerful convolutional neural networks. In our approach, we use a gradient descent method to iteratively modify an input image based on the pixel-wise gradients to reduce a cost function measuring the class-specific objectness of the image. The pixel-wise gradients can be efficiently computed using the back-propagation algorithm. The discrepancy between the modified image and the original one may be used  as a saliency map for the image. Moreover, we have further proposed several new training methods to learn saliency-specific convolutional nets for object saliency detection, in order to leverage the available pixel-wise segmentation information. Our methods are extremely computationally efficient (processing 20-40 images per second in one GPU). In this work, we use the computed saliency maps for image segmentation. Experimental results on two benchmark tasks, namely Microsoft COCO and Pascal VOC 2012, have shown that our proposed methods can generate high-quality salience maps, clearly outperforming many existing methods. In particular, our approaches excel in handling many difficult images, which contain complex background, highly-variable salient objects, multiple objects, and/or very small salient objects.
\end{abstract}

\section{Introduction}
\label{sec_introduction}

In the past few years, deep convolutional neural networks (DCNNs) \cite{lecun1995convolutional} have achieved the state of the art performance in many computer vision tasks, starting from image recognition \cite{NIPS2012_AlexNet,GoogleLeNet_2014,VGG_2015} and object localization \cite{overfeat_2014} and more recently extending to object detection and semantic image segmentation \cite{r-cnn_2014,hariharan2014simultaneous}.
These successes are largely attributed to the capacity that large-scale DCNNs can effectively learn end-to-end from a large amount of labelled images in a supervised learning mode.

In this paper, we consider to apply the popular deep learning techniques to another computer vision problem, namely object saliency detection.
The saliency detection attempts to locate the objects that have the most interests in an image, where human may also pay more attention on the image \cite{liu2011learning}.
The main goal of the saliency detection is to compute a saliency map that topographically represents the level of saliency
for visual attention \cite{BooleanMap-2013}.
For each pixel in an image, the saliency map can provide how likely this pixel belongs to the salient objects \cite{borji2012salient}.
Computing such saliency maps has recently raised a great amount of research interest \cite{visual-model-2013}.
The computed saliency maps have been shown to be beneficial to various vision tasks, such as
image segmentation \cite{cheng2011global}, object recognition  and visual tracking.
The saliency detection has been extensively studied in computer vision. A variety of methods have been proposed to generate the saliency maps for images. Under the assumption that the salient objects probably are the parts that significantly differ from their surroundings, most of the existing methods use low-level image features to detect saliency based on the criteria related to contrast, rarity and symmetry of image patches \cite{cheng2011global,liu2011learning,riche2012rare,borji2012salient}. In some cases, the global topological cues may be leveraged to refine the perceptual saliency maps \cite{harel2006graph,BooleanMap-2013,li2010probabilistic}. In these methods, the saliency is normally measured based on different mathematical models, including decision theoretic models, Bayesian models, information theoretic models, graphical models, spectral analysis models \cite{visual-model-2013}.

In this paper, we propose a novel deep learning method for the object saliency detection based on the powerful DCNNs. As shown in \cite{NIPS2012_AlexNet,GoogleLeNet_2014,VGG_2015}, relying on a well-trained DCNN, we can achieve a fairly high accuracy in object category recognition for many real-world images. Even though DCNNs can recognize what objects are contained in an image, it is not straightforward for DCNNs to precisely locate the recognized objects in the image. In \cite{overfeat_2014,r-cnn_2014,hariharan2014simultaneous}, some
rather complicated and time-consuming post-processing stages are needed to detect and locate the objects for semantic image segmentation. In this work, we propose a much simpler and more computationally efficient method to generate a class-specific object saliency map directly from the classification DCNN model. In our approach, we use a gradient descent (GD) method to iteratively modify each input image based on the pixel-wise gradients to reduce a cost function measuring the objectness of the image. The gradients with respect to all image pixels can be efficiently computed using the back-propagation algorithm for DCNNs. At the end, the discrepancy between the modified image and the original one is calculated as the saliency map for this image. Moreover, as more and more images with pixel-wise segmentation labels become available, e.g. \cite{Everingham10,lin2014microsoft}, we further propose two more methods to leverage the available pixel-wise segmentation information to learn saliency-specific DCNNs for the object saliency detection. In these methods, the original images as well as the corresponding masked images, in which all objects are masked out according to the pixel-wise labels, are used to train two DCNNs whose output labels are modified to include the masked objects and/or the original objects. Afterwards, we similarly use the GD method to modify each input image to reduce two cost functions formulated to measure the objectness for each case. The saliency map is generated in the same way as the discrepancy between the original and modified images. Since we only need to run a very small number of GD iterations in the saliency detection, our methods are extremely computationally efficient (processing 20-40 images per second in one GPU).
The computed saliency maps may be used for many computer vision tasks.
In this work, as one particular application, we use the computed saliency maps to drive an popular image segmenter in \cite{cArbelaez14} to perform image segmentation. Experimental results on two databases, namely Microsoft COCO \cite{lin2014microsoft} and Pascal VOC 2012 \cite{Everingham10}, have shown that our proposed methods can generate high-quality salience maps, clearly outperforming many existing methods. In particular, our DCNN-based approaches excel on many difficult images, containing complex background, highly-variable salient objects, multiple objects, and/or very small objects.

\section{Related Work}
\label{sec_related_work}

In the literature, the previous saliency detection methods mostly adopt the well-known bottom-up strategy \cite{cheng2011global,liu2011learning,riche2012rare,borji2012salient}. They relies on the local image features derived from patches to detect contrast, rarity and symmetry to identify the salient objects in an image. Meanwhile, some other methods have been proposed to take into account some global information or prior knowledge to screen the local features. For example,  in \cite{BooleanMap-2013}, a boolean map is created to represent global topological cues in an image, which in turn is used to guide the generation of saliency maps.
In \cite{li2010probabilistic}, the visual saliency algorithm considers the prior information and the local features simultaneously in a probabilistic model.
The algorithm defines task-related components as the prior information to help the feature selection procedure.
The traditional saliency detection methods normally work well for the images containing simple dominant foreground objects in homogenous backgrounds. However, they are usually not robust enough to handle images containing complex scenes \cite{li2014visual}. 

As an important application, the saliency maps may be used as a good guidance for various image segmentation algorithms. In \cite{donoser2009saliency}, a recursive segmentation process is used, where each iteration focuses on different saliency regions. As a result, the algorithm can output several potential segmentation candidates from the saliency maps. These candidates may be further merged by maximizing likelihood at all image pixels by considering the low-level features like colour and texture. In \cite{cheng2011global}, a region contrast based image saliency method is proposed to generate the saliency maps, and the SaliencyCut algorithm is used derive image segmentation from the saliency maps. The SaliencyCut algorithm is based on the standard GrabCut \cite{rother2004grabcut} but it uses the proposed saliency maps instead of manually selected bounding boxes for initialization.

Recently, some deep learning techniques have been proposed for object detection and semantic image segmentation \cite{overfeat_2014,r-cnn_2014,hariharan2014simultaneous}. These methods typically use DCNNs to examine a large number of region proposals from other algorithms, and use the features generated by DCNNs along with other post-stage classifiers to localize the target objects. They initially rely on bounding boxes for object detection. More recently,  more and more methods are proposed to directly generate pixel-wise image segmentation, e.g. \cite{hariharan2014simultaneous}.
In this paper, instead of directly generating the high-level semantic segmentation from DCNNs, we propose to use DCNNs to generate middle-level saliency maps in a very efficient way, which may be fed to other traditional computer vision algorithms for various vision tasks, such as semantic segmentation, video tracking, etc.

The work in \cite{Oxford-cnn-2014} is the most relevant to the work in this paper. In \cite{Oxford-cnn-2014}, the authors have borrowed the idea of explanation vectors in \cite{Baehrens-2010} to generate a static pixel-wise gradient vector of the network learning objective function, and use it as a saliency map. In our work, we instead use an iterative gradient descent method to generate more reliable and robust saliency maps. More importantly, we have proposed two new methods to learn saliency-specific DCNNs and define the corresponding cost functions, which measure objectness in each model for salinecy detection. 


\begin{figure*}[t]
\begin{center}
    \includegraphics[width=0.9\linewidth]{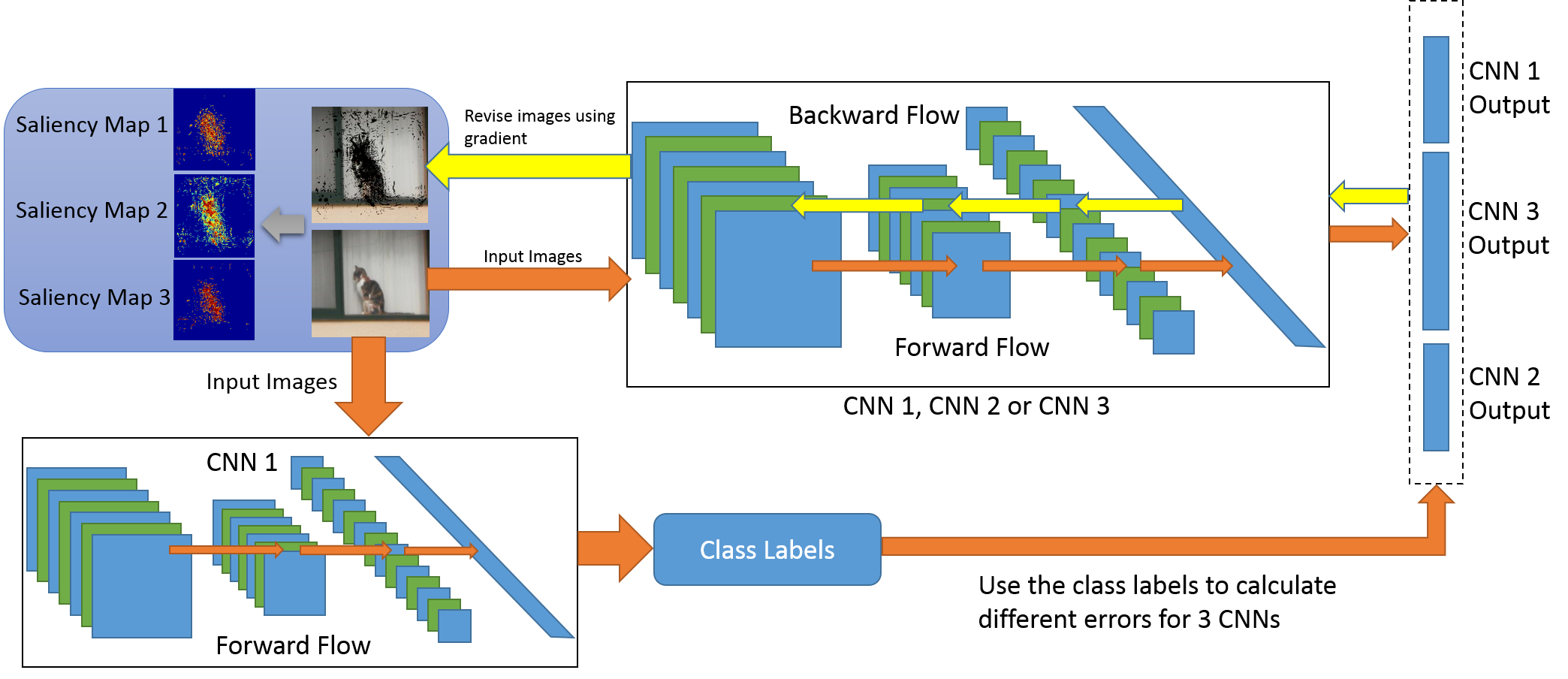}
\end{center}
\caption{The proposed method to generate the object-specific saliency maps directly from DCNNs.}
\label{Fig:Overview_saliency}
\end{figure*}

\section{Our Approach for Object Saliency Detection}
\label{sec_sal_seg}

As we have known, DCNNs can automatically learn all sorts of features from a large amount of labelled images, and a well-trained DCNN can achieve a very good classification accuracy in recognizing objects in images. In this work, based on the idea of explanation vectors in \cite{Baehrens-2010}, we argue that the classification DCNNs themselves may have learned enough features and information to generate good object saliency for the images. Extending a preliminary study in \cite{Oxford-cnn-2014}, we explore several novel methods to generate the saliency maps directly from DCNNs. The key idea of our approaches is shown in Figure~\ref{Fig:Overview_saliency}. After an input image is recognized by a DCNN as containing one particular object, if we can modify the input image in such a way that the DCNN no longer recognizes the object from it, the discrepancy between the modified image and the original one may serve as a good saliency map for the recognized object. In this paper, we propose to use a gradient descent (GD) method to iteratively modify the input image based on the pixel-wise gradients to reduce a cost function formulated in  the output layer of the DCNN to measure the class-specific objectness. The gradients are
computed by applying the back-propagation procedure all the way to the input layer.

In section~\ref{subsec_learning_saliency}, we first introduce several different ways to learn DCNNs for saliency detection. In section~\ref{subsec_generate_saliency}, we present our algorithm used to generate the saliency maps from DCNNs in detail.

%

\subsection{Learning DCNNs for Object Saliency}
\label{subsec_learning_saliency}

\begin{figure*}[t]
\begin{center}
    \includegraphics[width=0.9\linewidth]{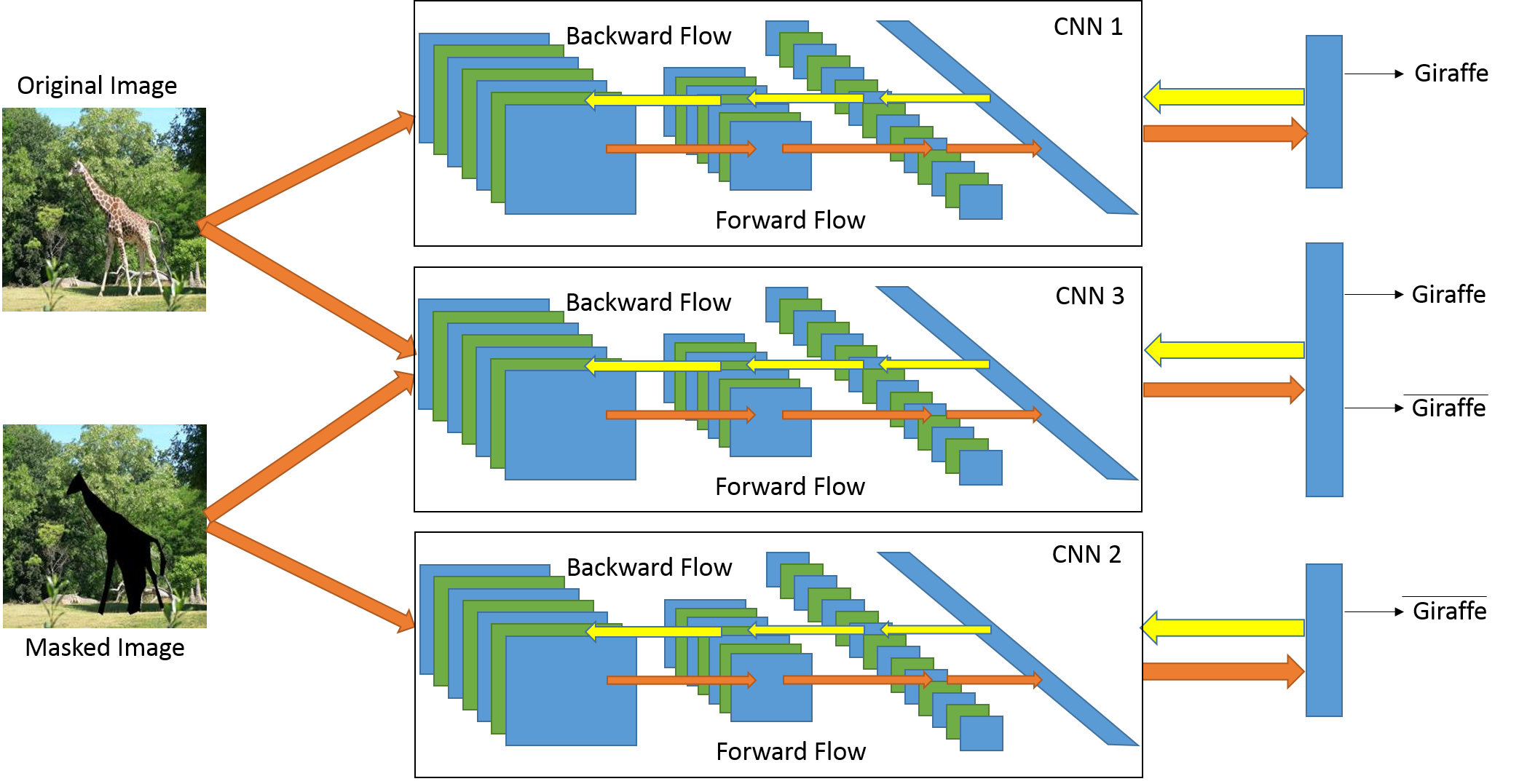}
\end{center}
\caption{The proposed training procedure to learn DCNNs for object saliency detection.}
\label{Fig:Overview_training}
\end{figure*}

Comparing with the traditional bottom-up methods, DCNNs may potentially learn more prior information for saliency detection. The first type is the class prior, which is provided by class labels of all training images. The second one is the pixel-wise object prior, which may be available as the object masking information in some data sets.

First of all, the regular classification DCNN may be used for saliency detection, which is named as CNN1 hereafter. As shown in Figure~\ref{Fig:Overview_training}, CNN1 takes an image as input and it contains a node in the output layer for each object category. CNNs is trained using all labeled images in the training set.

If the pixel-wise object masking information is available, we may mask out the corresponding objects in the original images to generate the so-called masked images. In this way, we may learn different DCNNs to learn the pixel-wise masking information, which will lead to much better DCNNs for the saliency detection purpose.
For example, we may learn another DCNN with the masked images only, named as CNN2. As in Figure ~\ref{Fig:Overview_training}, CNN2 is trained by using all masked images in the training set as input and it has a node in the output layer corresponding to each masked object class.

Moreover,  as shown in Figure ~\ref{Fig:Overview_training}, we train a slightly modified DCNN, named as CNN3, with both original labelled images as well as all masked images, in which all labelled objects are masked out based on the pixel-wise masking. For CNN3, we expand its output layer to include two nodes for each object category: one for the normal objects and the other for the masked objects. For example, when we use an original image containing a giraffe to learn CNN3, we use the label information corresponding to the regular giraffe node in the output layer, denoted as $\mbox{\it Giraffe}$. Meanwhile, when we use the same image with the animal region masked out, we use the label information corresponding to the masked giraffe node in the output layer, denoted as $\overline{\mbox{\it Giraffe}}$. Comparing with CNN2, CNN3 is trained in a way to learn the contrast information between original labelled images and their masked versions.

\subsection{Generating Saliency Maps from DCNNs}
\label{subsec_generate_saliency}

After the three DCNNs (CNN1, CNN2 and CNN3) are learned, we may apply our saliency detection methods to generate the class-specific object saliency map, as shown in Figure~\ref{Fig:Overview_saliency}.

For each input image, we firstly use CNN1 to generate its class label, denoted as $l$, as in a normal classification step.
Next, we may use one of the DCNNs to generate the saliency map. In this step, the selected DCNN is kept unchanged and instead we attempt to modify the input image in the pixel level to reduce a cost function, which is defined to measure the class-specific objectness in each case.
In the following, we introduce how to define the cost function for each DCNN and the details to generate the saliency maps.

For CNN1, we denote its output nodes after softmax as $\{ y^{(1)}_i \; | \; i=1,\cdots,N \}$, each of which corresponds to one class label ($N$ classes in total). Assume an input image $X$ is recognized as class $l$, we may define the following cost function to measure the class-specific objectness in this case:
\begin{equation}\label{eq-costfunction-CNN1}
{\cal F}^{(1)} (X | l ) = \ln y^{(1)}_l.
\end{equation}

The key idea here is that we try to modify the image $X$ to reduce the above cost function and hopefully the underlying object (belonging to class $l$) will be removed as the consequence. In this paper, we propose to use an iterative GD procedure to modify $X$ as follows:
\begin{equation}\label{eq-SGD-saliency}
X^{(t+1)} \gets X^{(t)} - \epsilon \cdot \max \left(\frac{\partial {\cal F}^{(1)} (X | l ) }{\partial X} \Big|_{X=X^{(t)}}, 0 \right)
\end{equation}
where $\epsilon$ is a learning rate, and we floor all negative gradients in the GD updates. 
We have observed in our experiments that the cost function ${\cal F}^{(1)} (X | l )$ can be significantly reduced by running only a small number of updates (typically 10-15 iterations) for each image.

We can easily compute the above gradients using the standard back-propagation algorithm. Based on the cost function ${\cal F}^{(1)}$ in eq.(\ref{eq-costfunction-CNN1}), we can derive the error signals in the output layer as
$e^{(1)}_i = \delta(i-l) - y^{(1)}_i$ ($i=1,\cdots, N$), where $\delta(\cdot)$ stands for the Kronecker delta function.
These error signals are back-propagated all the way to the input layer to derive the above gradient, $\frac{\partial {\cal F}^{(1)} (X | l ) }{\partial X}$, for saliency detection.

For CNN2, we denote its output nodes after softmax as $\{ y^{(2)}_i \; | \; i=1,\cdots,N \}$, each of which corresponds to one class of masked objects. Given an input image $X$ and its recognized class $l$ (from CNN1), we define the following cost function for this case:
\begin{equation}\label{eq-costfunction-CNN2}
{\cal F}^{(2)} (X | l ) = - \ln y^{(2)}_l.
\end{equation}

Similarly, we apply the above GD algorithm in eq.(\ref{eq-SGD-saliency}) to modify the image to reduce this cost function. By reducing ${\cal F}^{(2)}$, we try to increase the probability of the corresponding masked class. Intuitively, we attempt to alter the input image to match the masked images in that class as much as possible. In the same way, the error signals in the output layer can be simply derived as
$e^{(2)}_i =  y^{(2)}_i - \delta(i-l)$ ($i=1,\cdots, N$), which are back-propagated all the way to the input layer to compute
$\frac{\partial {\cal F}^{(2)} (X | l ) }{\partial X}$.

Finally, for CNN3, we denote its output nodes after softmax as $\{ y^{(3)}_i \; | \; i=1,\cdots,2N \}$, each of which corresponds to either an image class or a masked class. Given an input input image $X$ and its recognized class $l$, we find the output node corresponding to the masked class of $l$, denoted as $\bar{l}$. We define the cost function for CNN3 as follows:
\begin{equation}\label{eq-costfunction-CNN3}
{\cal F}^{(3)} (X | l ) = - \ln y^{(3)}_{\bar{l}}.
\end{equation}

Similarly, the image is modified by running the GD algorithm in eq.(\ref{eq-SGD-saliency})
to reduce ${\cal F}^{(3)}$, or equivalently increase $y^{(3)}_{\bar{l}}$.
 Since all output nodes are normalized by softmax, by increasing $y^{(3)}_{\bar{l}}$, its original output node $y^{(3)}_{l}$ will be reduced accordingly. Intuitively speaking, by doing so,  we attempt to use the contrast information learned by CNN3 to modify an image from its original class to match the corresponding masked version for the object saliency detection.
Similarly, the error signals in the output layer is derived as
$e^{(3)}_i =  y^{(3)}_{i} - \delta(i-\bar{l})$, where $i=1,\cdots, 2N$.

At the end of the gradient descent updates, the object saliency map is computed as the difference between the modified image and the original one, i.e. $X^{(0)}-X^{(T)}$. For colour images, we average the differences over the RGB channels to obtain a pixel-wise raw saliency map, which is then normalized to be of unit norm. After that, we may apply a simple threshold to filter out some background noises of the raw saliency maps.
The entire algorithm to generate the raw saliency maps is shown in {\bf Algorithm 1}. 

\begin{algorithm}[tb]
   \caption{GD based Object Saliency Detection}
   \label{alg:Saliency}
\begin{algorithmic}
   \STATE {\bfseries Input:} an input image $X$, CNN1, CNN2 and CNN3;

   \STATE Use CNN1 to recognize the object label for $X$ as $l$;

   \STATE Choose a saliency model (CNN1 or CNN2 or CNN3);

   \STATE $X^{(0)}=X$;

   \FOR{{\bfseries each} epoch $t=1$ {\bfseries to} $T$}

   \STATE {\bfseries forward pass}: compute the cost function ${\cal F}(X|l)$ ;

   \STATE {\bfseries backward pass}: back-propagate to input layer to compute gradient: $\frac{\partial {\cal F}(X|l)}{\partial X}$;

   \STATE $X^{(t)} \gets X^{(t-1)} - \epsilon \cdot \max\left(\frac{\partial {\cal F}(X|l)}{\partial X},0 \right)$;

   \ENDFOR

   \STATE Average over RGB: ${\bf S}= \frac{1}{3} \sum_{i=1}^3 (X^{(0)}_i - X^{(T)}_i )$;

   \STATE Prune noises with a threshold $\theta$: ${\bf S} = \max({\bf S}-\theta,0)$;

   \STATE Normalize: ${\bf S} = \frac{\bf S}{\| {\bf S} \|}$;


   \STATE {\bfseries Output:} the raw saliency map ${\bf S}$;

\end{algorithmic}
\end{algorithm}

For each image, we can obtain 3 different saliency maps with the three different DCNNs. We have found that we may obtain even better results if we combine the saliency maps from CNN2 and CNN3 by taking an average between them. We can also use a simple image dilation and erosion method to smooth the raw saliency maps to derive the final saliency maps.

\section{Saliency Refinement and Image Segmentation}
\label{sec_image_segmentation}

Here, as one application, we use the derived saliency maps to perform semantic image segmentation. 

Inspired by the recent work in ~\cite{hariharan2014simultaneous}, we aim to refine our saliency map using segmentation and also achieve a binary salient object segmentation. We make use of a recent state-of-art image segmentation tool called Multiscale Combinatorial Grouping (MCG)~\cite{cArbelaez14}, which provides us with a well-defined contour map and also a set of object proposals. The idea of refining the saliency map is simple: we randomly select $50$ points from salient point sets and use these selected points as seed information to perform an interactive image segmentation. We restrict it to be a binary segmentation to extract salient foreground. We independently run this experiment $100$ times and average the binary segmentation results, then we can get a refined saliency. 

To obtain the final binary salient object segmentation, we use the top $50$ object proposals generated by MCG. For each proposal associated with super-pixel segmentation, we choose the one with the highest Jaccard index value with a thresholded binary mask from the provided saliency map. Specifically, given the final saliency map as $\mathbf{S}$, we get a binary mask $\mathbf{\mathcal{M}}_1 = I\{\mathbf{S}>\delta\}$, where $\delta$ is a threshold (we set it to be $0.5$ in this work). For each super-pixel segmentation from each proposal, denoted as $\mathbf{\mathcal{M}}_2$, we calculate the Jaccard index as follows:
$$
Jaccard(\mathbf{\mathcal{M}}_1, \mathbf{\mathcal{M}}_2) = \frac{\|\mathbf{\mathcal{M}}_1 \bigcap \mathbf{\mathcal{M}}_2\|}{\|\mathbf{\mathcal{M}}_1 \bigcup \mathbf{\mathcal{M}}_2\|}
$$
The super-pixel segmentation that has the largest Jaccard index with the thresholded saliency map is chosen as the final salient object segmentation.

\section{Experiments}
\label{sec_experiments}

We select two benchmark databases to evaluate the performance of the proposed object saliency detection and image segmentation methods, namely Microsoft COCO \cite{lin2014microsoft} and Pascal VOC 2012 \cite{Everingham10}. Both databases provide the class label of each image as well as the pixel-wise segmentation map (ground truth), thus we can generate the masked images to train the required DCNNs in our propsed methods. Here we compare our approaches with two exisiting methods: i) the first one is the Region Contrast saliency method and the SaliencyCut segmentation method in \cite{cheng2011global}. This method is one of the most popular  bottom-up image saliency detection methods in the literature and it has achieved the state-of-the-art image saliency and segmentation performance on many tasks; ii) the second one is the DCNN based image saliency detection method proposed in \cite{Oxford-cnn-2014}. Similar to our approaches, this method also use DCNNs and the back-propagation algorithm to generate saliency maps. In our experiments, we use the precision-recall curves (PR-curves) against the ground truth as one metric to evaluate the performance of saliency detection.
As \cite{cheng2011global}, for each saliency map, we vary the cutoff threshold from $0$ to $255$ to generate $256$ precision and recall pairs, which are used to plot a PR-curve. Besides,
we  also use $F_{\beta}$ to measure the performance for both saliency detection and segmentation, which is  calculated based on precision $Prec$ and recall $Rec$ values with a non-negative weight parameter $\beta$ as follows \cite{borji2012salient}:
\begin{equation}
\label{eq-Fbeta}
F_{\beta} = \frac{(1+{\beta}^2)Prec \times Rec}{{\beta}^2 Prec + Rec}
\end{equation}

In this paper, we follow \cite{cheng2011global} to set ${\beta}^2 = 0.3$ to emphasize the importance of $Prec$. Note that we only get a single $F_{\beta}$ value for each binary segmentation map for segmentation. However,
we may derive a sequence of $F_{\beta}$ values along the PR-curve for each saliency map and the largest one is selected as the performance measure (see \cite{borji2012salient}).

\subsection{Databases}

Microsoft COCO \cite{lin2014microsoft} is a new image database that may be used for several vision tasks including image classification and segmentation. The database currently contains $82,783$ training images and $40,504$ validation images with $80$ labeled categories. In our experiments, we only select the images that contain one category of objects because these images are more compatible with the available DCNN baseline, which is normally trained using the ImageNet data. The selected COCO subset contains $6869$ training images and $3479$ validation images with $18$ different classes. 

Pascal VOC 2012 database \cite{Everingham10} can also be used for our proposed algorithms, but its sample size is much smaller comparing with COCO. We use the whole dataset, which has $1464$ training images and $1449$ validation images with $20$ label categories in total. For images that are labelled to have more than one class of objects, we use the area of the labelled objects to measure their importance and use the class of the most important object to label the images for our DCNN training process.

As we have mentioned earlier, we need to train the three DCNNs, i.e., CNN1, CNN2 and CNN3, for each dataset. However,  because the training sets are relatively small in both COCO and Pascal, we have used a well-trained DCNN for the ImageNet database, which contains 5 convolutional layers and 2 fully connected layers\footnote{We use the net {\em imagenet-vgg-s} in http://www.vlfeat.org/matconvnet/ \cite{Chatfield14}.}.  We only use the above-mentioned training data to fine-tune this DCNN for each task with MatConvNet in \cite{MatConvNet-2014}. For the Pascal VOC 2012 data, we further use 5-fold cross-validation to expand the training sample size. We use the training set and about 80\% of the validation data to fine-tune the model and it is used to test the remaining 20\% of data. We rotate five times to cover the entire test set. In Table~\ref{table:CNNbaseline}, we have listed the top-1 and top-5 classification  error rates when the fine-tuned DCNNs are used to recognize the test sets on these two tasks.

\begin{table}
\begin{center}
\begin{tabular}{|c|c|c|c|c|c|r|}
\hline
               &         & CNN1     & CNN2     & CNN3    \\
\hline\hline
MS  & Top-1     & 12.2\%   & 19.1\%   & 16.7\%   \\
 COCO      & Top-5     & 2.4\%    & 3.2\%    & 4.0\%    \\ \hline
Pascal & Top-1 & 20.3\%   & 35.1\%   & 26.5\%   \\
VOC 2012 & Top-5 & 3.1\%    & 8.4\%    & 9.7\%   \\
\hline
\end{tabular}
\end{center}
\caption{The classification error rates of three CNNs on the MS COCO and Pascal VOC 2012 test sets.}
\label{table:CNNbaseline}
\end{table}

The classification errors on the test sets imply that the training sample size is still not enough for training deep convolutional networks well, especially for Pascal VOC 2012. However, as we will see, the proposed algorithms can still yield good performance for saliency detection and segmentation. If we have more training data that include class labels and the masked images, we may expect even better saliency and segmentation results.

\subsection{Saliency and Segmentation Results}

In this part we will provide saliency detection and segmentation results on these two databases. In the following,  the PR-curves, $F_{\beta}$ values and some sample images will be used to compare different methods.

\subsubsection{Microsoft COCO}

\begin{figure}[htb]
\begin{center}
    \includegraphics[width=0.75\linewidth]{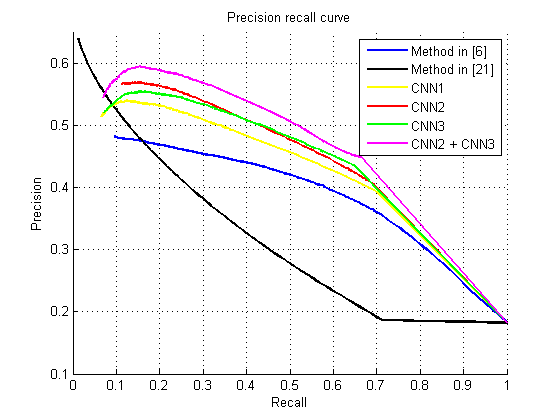}
\end{center}
\caption{The PR-curves of different saliency methods on the MS COCO test set.}
\label{Fig:PRCurve_COCO}
\end{figure}

For the object saliency detection, we first plot the PR-curves for different methods, which are all shown in Fig.~\ref{Fig:PRCurve_COCO}. From the PR-curves,  we can see that the performance of our proposed saliency detection methods significantly outperform the region contrast in \cite{cheng2011global} and the DCNN based saliency method in \cite{Oxford-cnn-2014}.  Moreover, it has shown that
CNN2 and CNN3 yields better performance than CNN1, which demonstrates that
the utilization of masked images in model training can further improve the saliency detection performance.

\begin{figure}[htb]
\begin{center}
    \includegraphics[width=0.75\linewidth]{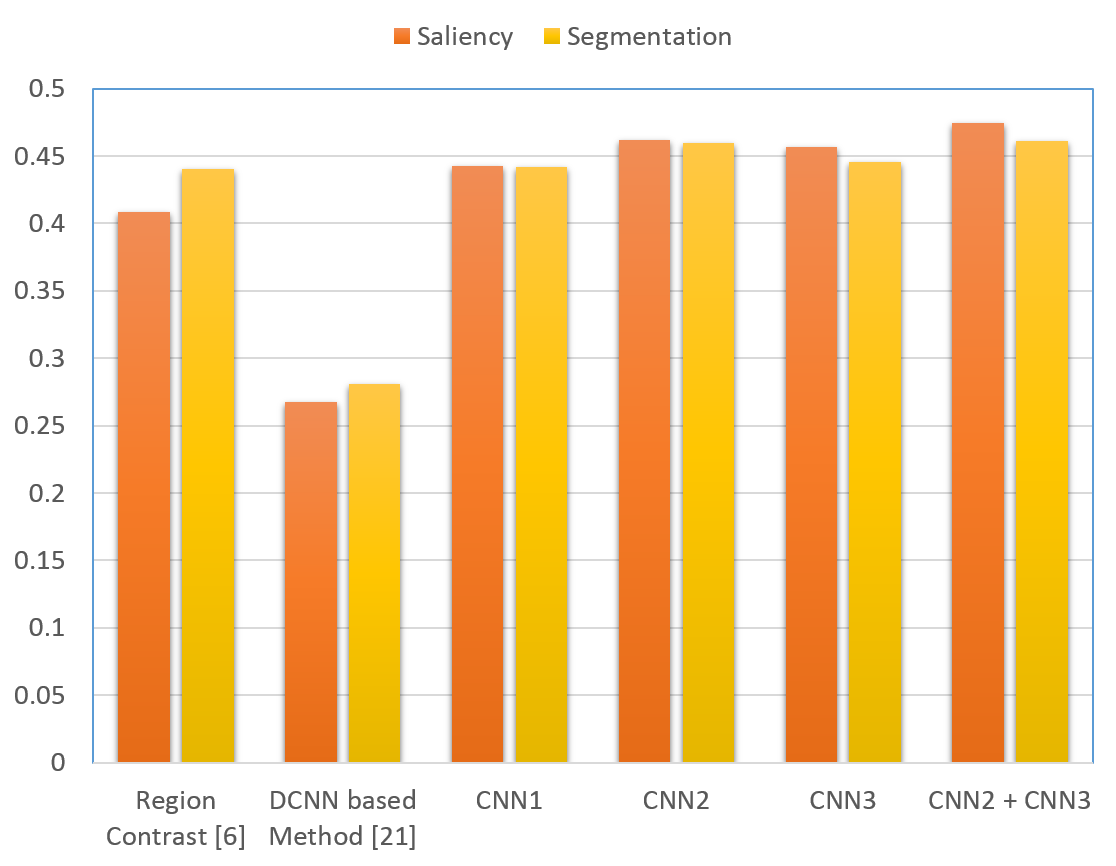}
\end{center}
\caption{The $F_{\beta}$ values of different saliency and segmentation methods on MS COCO test set.}
\label{Fig:F_COCO}
\end{figure}

Figure~\ref{Fig:F_COCO} shows the $F_{\beta}$ values of the different saliency and segmentation methods, from which we can see that the proposed three saliency detection methods give the better $F_{\beta}$ values than \cite{cheng2011global} and \cite{Oxford-cnn-2014}. Starting from our saliency maps, the MCG-based segmentation algorithm can yield a good performance as well. Moreover, the segmentation results have also shown the benefits to use the masked images as prior information in the DCNN training. Finally, in Figure~\ref{Fig:Saliency} (Column 1 to 5), we also provide some examples of the saliency detection and segmentation results from the COCO test set. From these examples we can see that the region contrast algorithm does not work well when the input images have complex background or contain highly variable salient objects, and this problem is fairly common among most bottom-up saliency and segmentation algorithms. On the other hand, we can also see that with the help of masked images in training our proposed DCNN-based saliency detection methods concentrate much better on the salient objects. Note that the segmentation results based on \cite{Oxford-cnn-2014} are not shown in Figure~\ref{Fig:Saliency} since they are significantly worse than others.

\subsubsection{Pascal VOC 2012}

\begin{figure}[htb]
\begin{center}
    \includegraphics[width=0.75\linewidth]{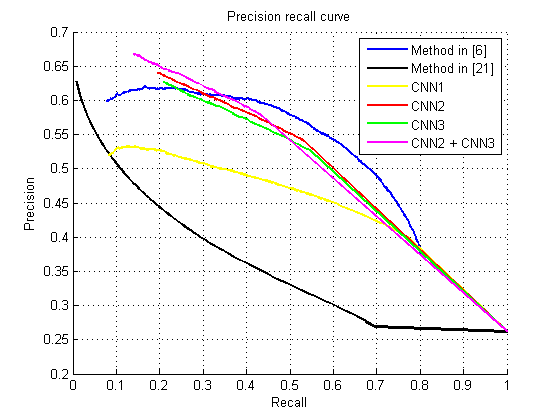}
\end{center}
\caption{The PR-curves of different saliency methods on Pascal VOC 2012 test set.}
\label{Fig:PRCurve_Pascal}
\end{figure}

\begin{figure}[htb]
\begin{center}
    \includegraphics[width=0.75\linewidth]{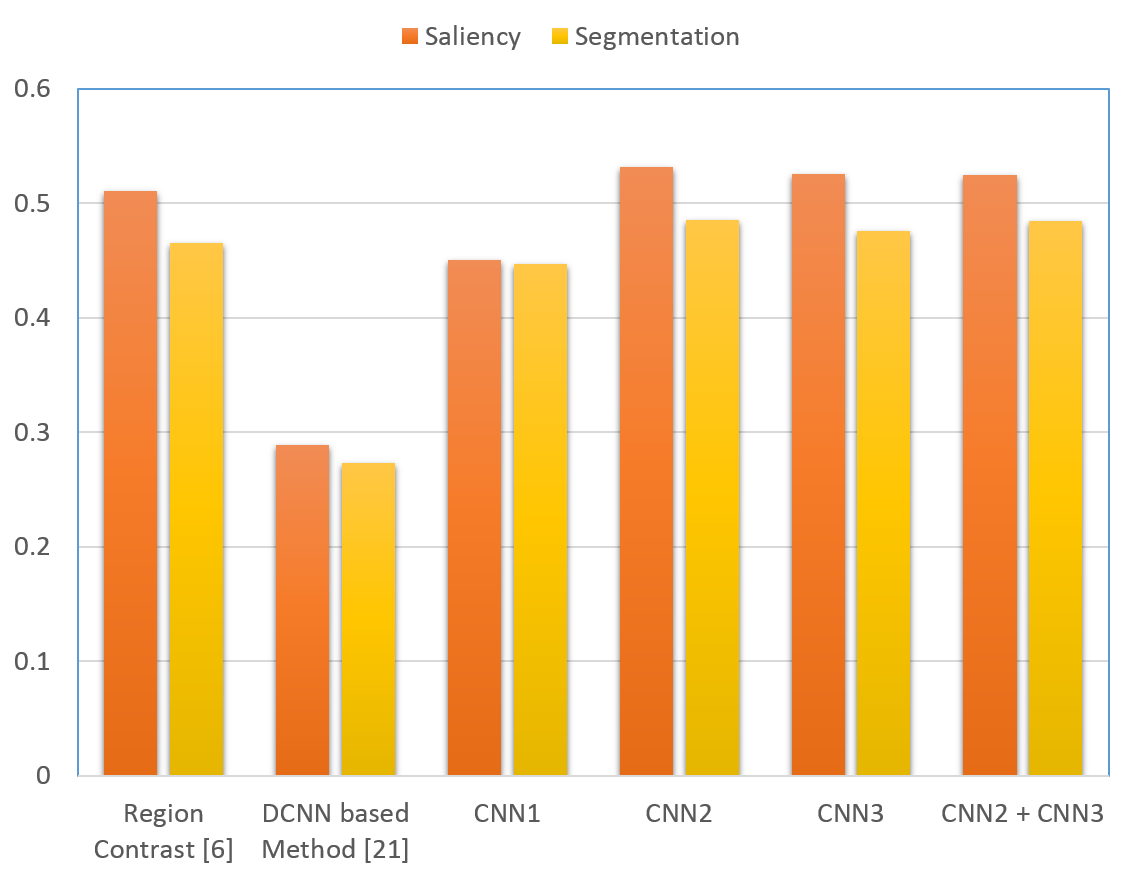}
\end{center}
\caption{The $F_{\beta}$ values of different saliency and segmentation methods on Pascal VOC 2012 test set.}
\label{Fig:F_Pascal}
\end{figure}

\begin{figure*}[tb]
\begin{minipage}{0.06\linewidth}
    \centerline{\scalebox{0.55}{(A) Original}}
\end{minipage}
\hfill
\begin{minipage}{0.06\linewidth}
    \centerline{\includegraphics[width=1.6cm]{890_im.jpg}}
\end{minipage}
\hfill
\begin{minipage}{0.06\linewidth}
    \centerline{\includegraphics[width=1.6cm]{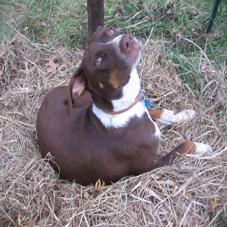}}
\end{minipage}
\hfill
\begin{minipage}{0.06\linewidth}
    \centerline{\includegraphics[width=1.6cm]{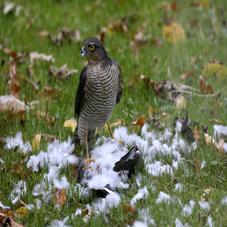}}
\end{minipage}
\hfill
\begin{minipage}{0.06\linewidth}
    \centerline{\includegraphics[width=1.6cm]{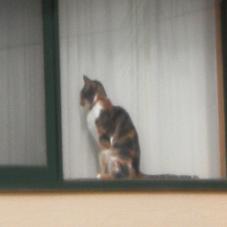}}
\end{minipage}
\hfill
\begin{minipage}{0.06\linewidth}
    \centerline{\includegraphics[width=1.6cm]{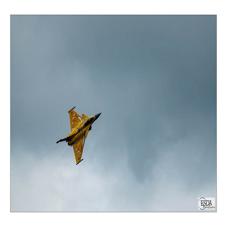}}
\end{minipage}
\hfill
\begin{minipage}{0.06\linewidth}
    \centerline{\includegraphics[width=1.6cm]{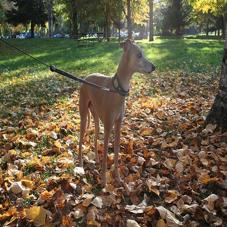}}
\end{minipage}
\hfill
\begin{minipage}{0.06\linewidth}
    \centerline{\includegraphics[width=1.6cm]{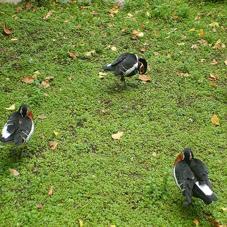}}
\end{minipage}
\hfill
\begin{minipage}{0.06\linewidth}
    \centerline{\includegraphics[width=1.6cm]{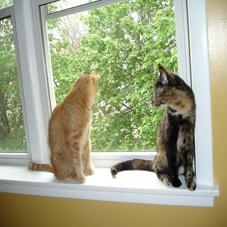}}
\end{minipage}
\hfill
\begin{minipage}{0.06\linewidth}
    \centerline{\includegraphics[width=1.6cm]{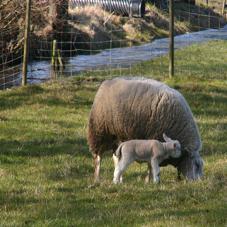}}
\end{minipage}
\hfill
\begin{minipage}{0.06\linewidth}
    \centerline{\includegraphics[width=1.6cm]{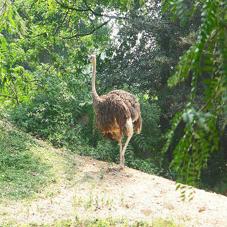}}
\end{minipage}
\vfill
\begin{minipage}{0.06\linewidth}
    \centerline{\scalebox{0.55}{(B) Ground truth}}
\end{minipage}
\hfill
\begin{minipage}{0.06\linewidth}
    \centerline{\includegraphics[width=1.6cm]{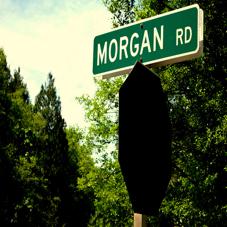}}
\end{minipage}
\hfill
\begin{minipage}{0.06\linewidth}
    \centerline{\includegraphics[width=1.6cm]{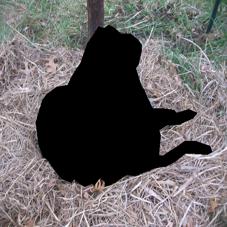}}
\end{minipage}
\hfill
\begin{minipage}{0.06\linewidth}
    \centerline{\includegraphics[width=1.6cm]{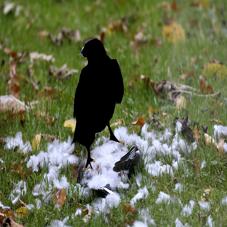}}
\end{minipage}
\hfill
\begin{minipage}{0.06\linewidth}
    \centerline{\includegraphics[width=1.6cm]{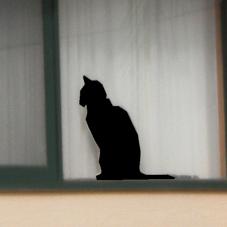}}
\end{minipage}
\hfill
\begin{minipage}{0.06\linewidth}
    \centerline{\includegraphics[width=1.6cm]{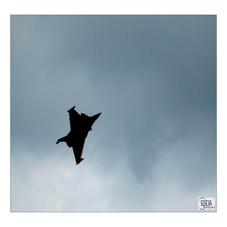}}
\end{minipage}
\hfill
\begin{minipage}{0.06\linewidth}
    \centerline{\includegraphics[width=1.6cm]{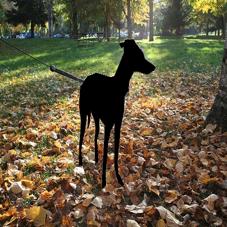}}
\end{minipage}
\hfill
\begin{minipage}{0.06\linewidth}
    \centerline{\includegraphics[width=1.6cm]{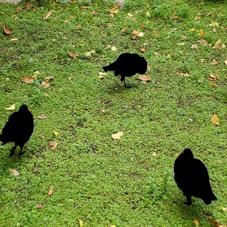}}
\end{minipage}
\hfill
\begin{minipage}{0.06\linewidth}
    \centerline{\includegraphics[width=1.6cm]{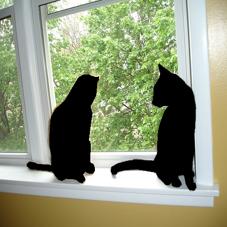}}
\end{minipage}
\hfill
\begin{minipage}{0.06\linewidth}
    \centerline{\includegraphics[width=1.6cm]{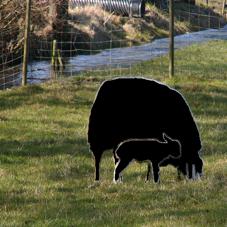}}
\end{minipage}
\hfill
\begin{minipage}{0.06\linewidth}
    \centerline{\includegraphics[width=1.6cm]{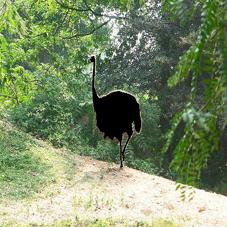}}
\end{minipage}
\vfill
\begin{minipage}{0.06\linewidth}
    \centerline{\scalebox{0.55}{(C) Region Contrast}}
    \centerline{\scalebox{0.55}{Saliency \cite{cheng2011global}}}
\end{minipage}
\hfill
\begin{minipage}{0.06\linewidth}
    \centerline{\includegraphics[width=1.6cm]{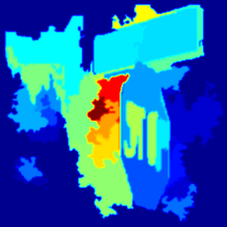}}
\end{minipage}
\hfill
\begin{minipage}{0.06\linewidth}
    \centerline{\includegraphics[width=1.6cm]{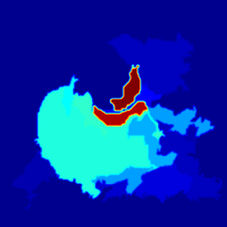}}
\end{minipage}
\hfill
\begin{minipage}{0.06\linewidth}
    \centerline{\includegraphics[width=1.6cm]{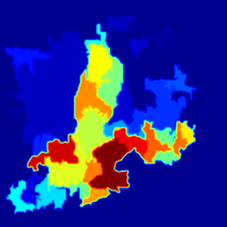}}
\end{minipage}
\hfill
\begin{minipage}{0.06\linewidth}
    \centerline{\includegraphics[width=1.6cm]{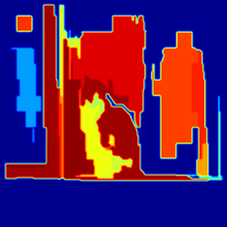}}
\end{minipage}
\hfill
\begin{minipage}{0.06\linewidth}
    \centerline{\includegraphics[width=1.6cm]{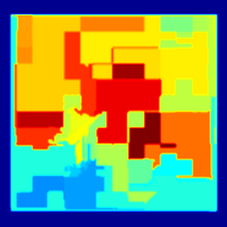}}
\end{minipage}
\hfill
\begin{minipage}{0.06\linewidth}
    \centerline{\includegraphics[width=1.6cm]{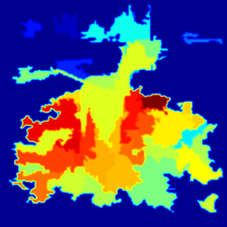}}
\end{minipage}
\hfill
\begin{minipage}{0.06\linewidth}
    \centerline{\includegraphics[width=1.6cm]{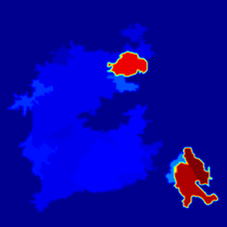}}
\end{minipage}
\hfill
\begin{minipage}{0.06\linewidth}
    \centerline{\includegraphics[width=1.6cm]{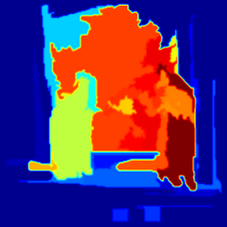}}
\end{minipage}
\hfill
\begin{minipage}{0.06\linewidth}
    \centerline{\includegraphics[width=1.6cm]{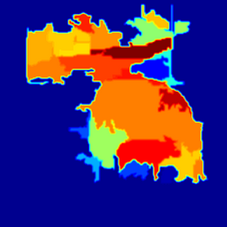}}
\end{minipage}
\hfill
\begin{minipage}{0.06\linewidth}
    \centerline{\includegraphics[width=1.6cm]{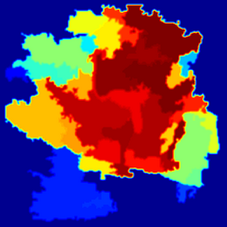}}
\end{minipage}
\vfill
\begin{minipage}{0.06\linewidth}
    \centerline{\scalebox{0.55}{(D) DCNN based}}
    \centerline{\scalebox{0.55}{method in \cite{Oxford-cnn-2014}}}
\end{minipage}
\hfill
\begin{minipage}{0.06\linewidth}
    \centerline{\includegraphics[width=1.6cm]{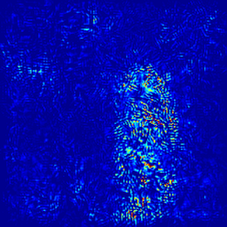}}
\end{minipage}
\hfill
\begin{minipage}{0.06\linewidth}
    \centerline{\includegraphics[width=1.6cm]{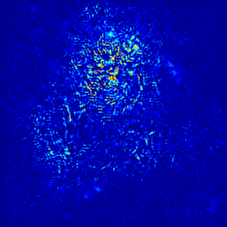}}
\end{minipage}
\hfill
\begin{minipage}{0.06\linewidth}
    \centerline{\includegraphics[width=1.6cm]{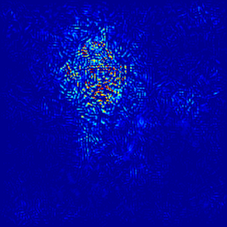}}
\end{minipage}
\hfill
\begin{minipage}{0.06\linewidth}
    \centerline{\includegraphics[width=1.6cm]{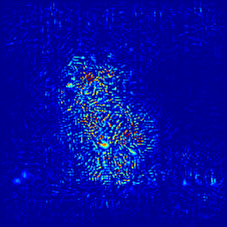}}
\end{minipage}
\hfill
\begin{minipage}{0.06\linewidth}
    \centerline{\includegraphics[width=1.6cm]{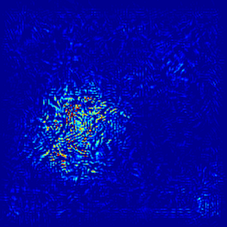}}
\end{minipage}
\hfill
\begin{minipage}{0.06\linewidth}
    \centerline{\includegraphics[width=1.6cm]{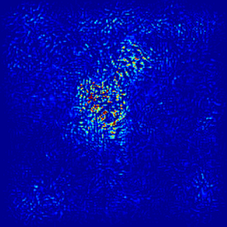}}
\end{minipage}
\hfill
\begin{minipage}{0.06\linewidth}
    \centerline{\includegraphics[width=1.6cm]{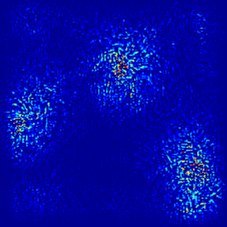}}
\end{minipage}
\hfill
\begin{minipage}{0.06\linewidth}
    \centerline{\includegraphics[width=1.6cm]{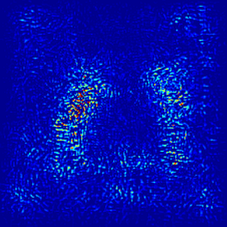}}
\end{minipage}
\hfill
\begin{minipage}{0.06\linewidth}
    \centerline{\includegraphics[width=1.6cm]{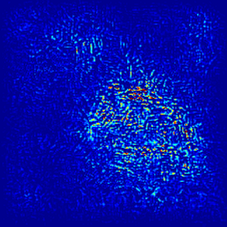}}
\end{minipage}
\hfill
\begin{minipage}{0.06\linewidth}
    \centerline{\includegraphics[width=1.6cm]{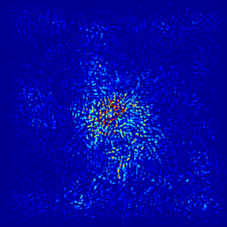}}
\end{minipage}
\vfill
\begin{minipage}{0.06\linewidth}
    \centerline{\scalebox{0.55}{(E) Raw saliency}}
    \centerline{\scalebox{0.55}{maps (CNN1)}}
\end{minipage}
\hfill
\begin{minipage}{0.06\linewidth}
    \centerline{\includegraphics[width=1.6cm]{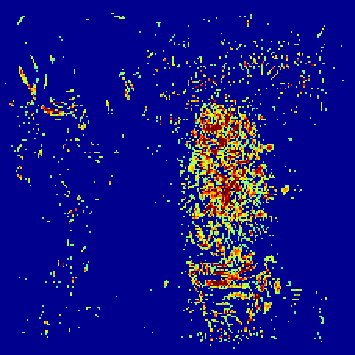}}
\end{minipage}
\hfill
\begin{minipage}{0.06\linewidth}
    \centerline{\includegraphics[width=1.6cm]{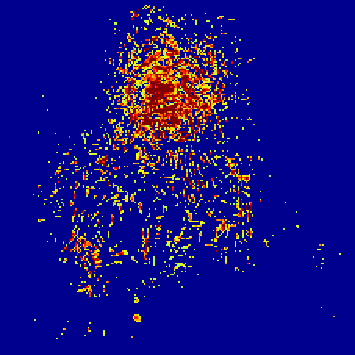}}
\end{minipage}
\hfill
\begin{minipage}{0.06\linewidth}
    \centerline{\includegraphics[width=1.6cm]{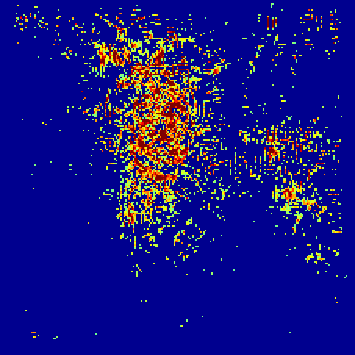}}
\end{minipage}
\hfill
\begin{minipage}{0.06\linewidth}
    \centerline{\includegraphics[width=1.6cm]{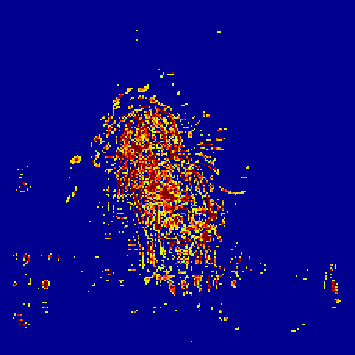}}
\end{minipage}
\hfill
\begin{minipage}{0.06\linewidth}
    \centerline{\includegraphics[width=1.6cm]{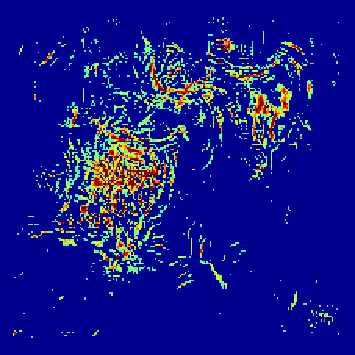}}
\end{minipage}
\hfill
\begin{minipage}{0.06\linewidth}
    \centerline{\includegraphics[width=1.6cm]{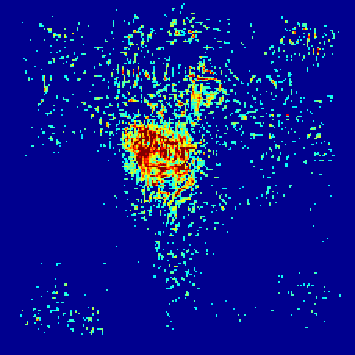}}
\end{minipage}
\hfill
\begin{minipage}{0.06\linewidth}
    \centerline{\includegraphics[width=1.6cm]{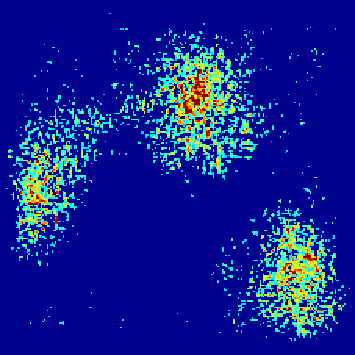}}
\end{minipage}
\hfill
\begin{minipage}{0.06\linewidth}
    \centerline{\includegraphics[width=1.6cm]{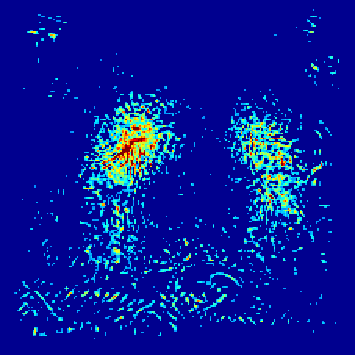}}
\end{minipage}
\hfill
\begin{minipage}{0.06\linewidth}
    \centerline{\includegraphics[width=1.6cm]{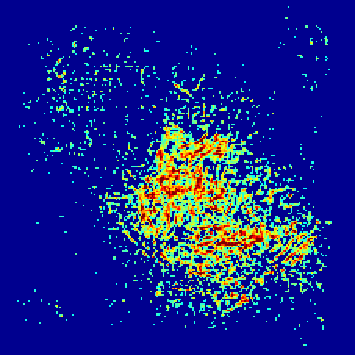}}
\end{minipage}
\hfill
\begin{minipage}{0.06\linewidth}
    \centerline{\includegraphics[width=1.6cm]{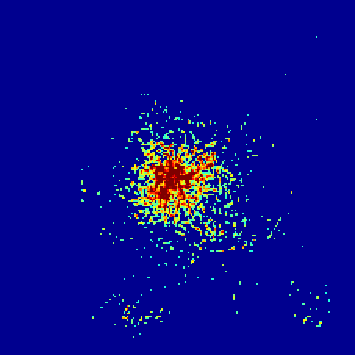}}
\end{minipage}
\vfill
\begin{minipage}{0.06\linewidth}
    \centerline{\scalebox{0.55}{(F) Raw saliency}}
    \centerline{\scalebox{0.55}{maps (CNN2)}}
\end{minipage}
\hfill
\begin{minipage}{0.06\linewidth}
    \centerline{\includegraphics[width=1.6cm]{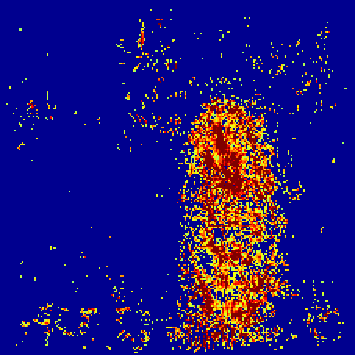}}
\end{minipage}
\hfill
\begin{minipage}{0.06\linewidth}
    \centerline{\includegraphics[width=1.6cm]{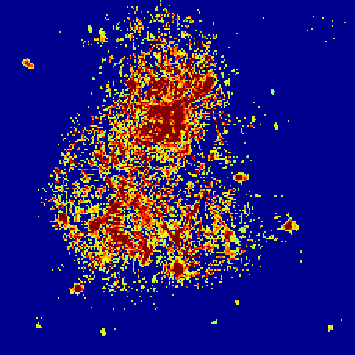}}
\end{minipage}
\hfill
\begin{minipage}{0.06\linewidth}
    \centerline{\includegraphics[width=1.6cm]{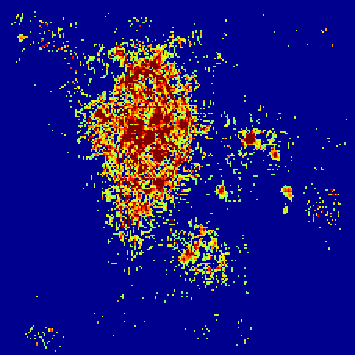}}
\end{minipage}
\hfill
\begin{minipage}{0.06\linewidth}
    \centerline{\includegraphics[width=1.6cm]{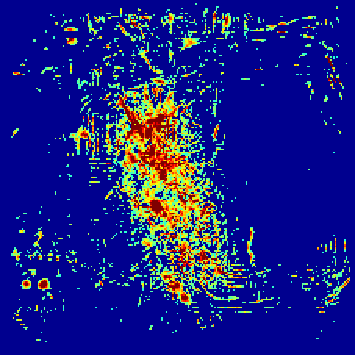}}
\end{minipage}
\hfill
\begin{minipage}{0.06\linewidth}
    \centerline{\includegraphics[width=1.6cm]{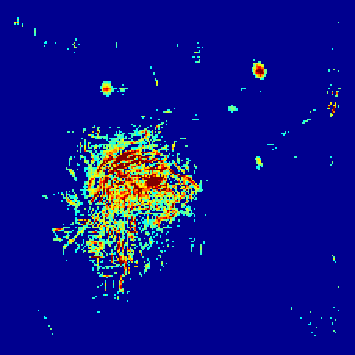}}
\end{minipage}
\hfill
\begin{minipage}{0.06\linewidth}
    \centerline{\includegraphics[width=1.6cm]{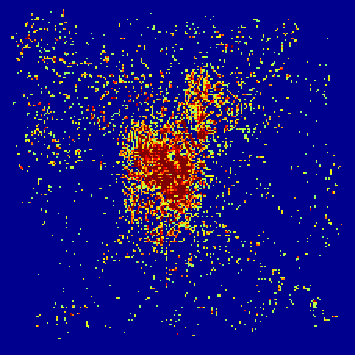}}
\end{minipage}
\hfill
\begin{minipage}{0.06\linewidth}
    \centerline{\includegraphics[width=1.6cm]{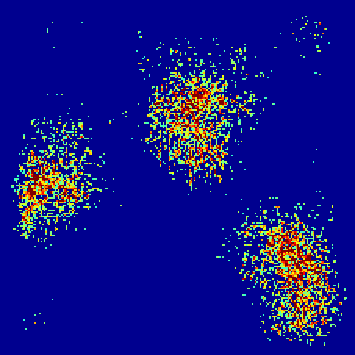}}
\end{minipage}
\hfill
\begin{minipage}{0.06\linewidth}
    \centerline{\includegraphics[width=1.6cm]{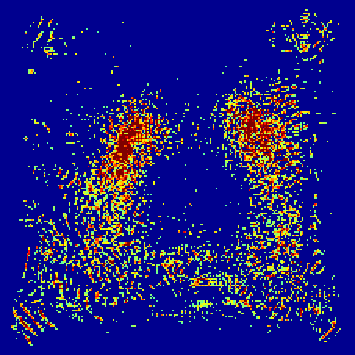}}
\end{minipage}
\hfill
\begin{minipage}{0.06\linewidth}
    \centerline{\includegraphics[width=1.6cm]{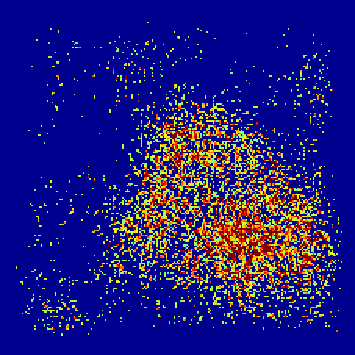}}
\end{minipage}
\hfill
\begin{minipage}{0.06\linewidth}
    \centerline{\includegraphics[width=1.6cm]{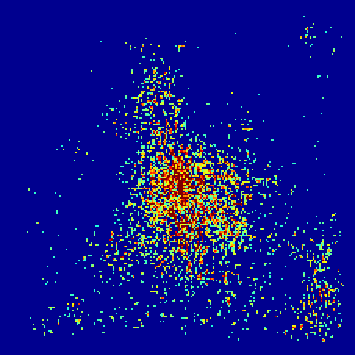}}
\end{minipage}
\vfill
\begin{minipage}{0.06\linewidth}
    \centerline{\scalebox{0.55}{(G) Raw saliency}}
    \centerline{\scalebox{0.55}{maps (CNN3)}}
\end{minipage}
\hfill
\begin{minipage}{0.06\linewidth}
    \centerline{\includegraphics[width=1.6cm]{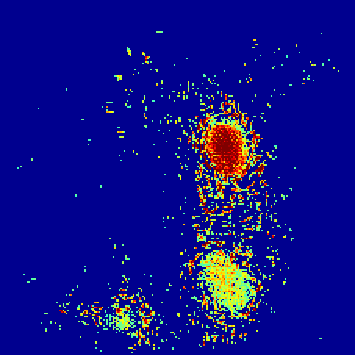}}
\end{minipage}
\hfill
\begin{minipage}{0.06\linewidth}
    \centerline{\includegraphics[width=1.6cm]{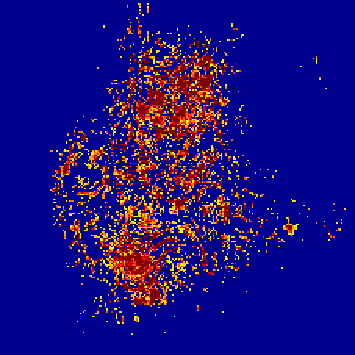}}
\end{minipage}
\hfill
\begin{minipage}{0.06\linewidth}
    \centerline{\includegraphics[width=1.6cm]{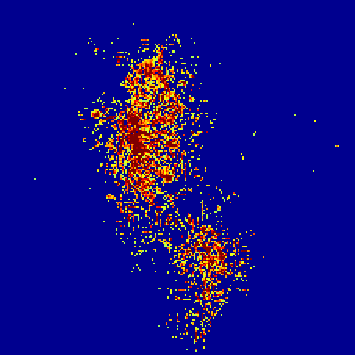}}
\end{minipage}
\hfill
\begin{minipage}{0.06\linewidth}
    \centerline{\includegraphics[width=1.6cm]{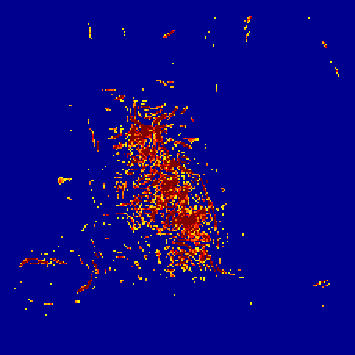}}
\end{minipage}
\hfill
\begin{minipage}{0.06\linewidth}
    \centerline{\includegraphics[width=1.6cm]{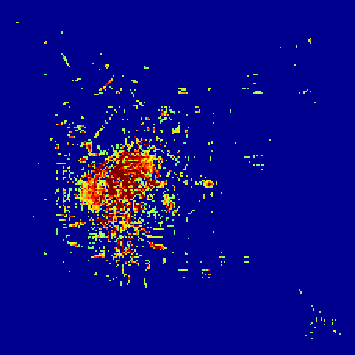}}
\end{minipage}
\hfill
\begin{minipage}{0.06\linewidth}
    \centerline{\includegraphics[width=1.6cm]{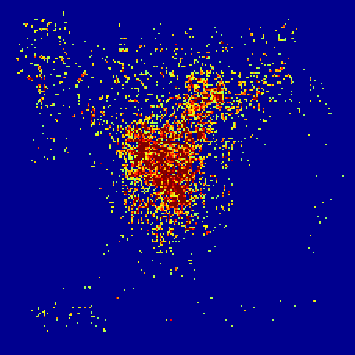}}
\end{minipage}
\hfill
\begin{minipage}{0.06\linewidth}
    \centerline{\includegraphics[width=1.6cm]{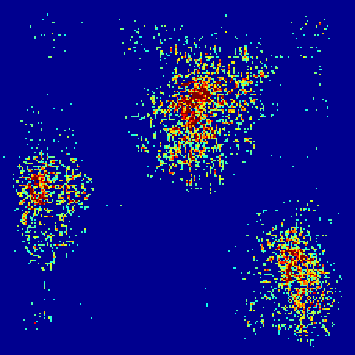}}
\end{minipage}
\hfill
\begin{minipage}{0.06\linewidth}
    \centerline{\includegraphics[width=1.6cm]{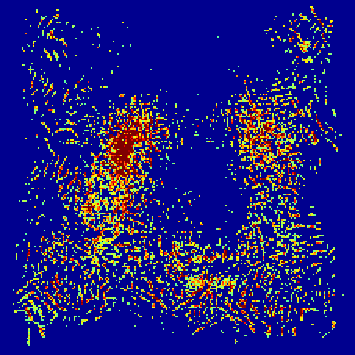}}
\end{minipage}
\hfill
\begin{minipage}{0.06\linewidth}
    \centerline{\includegraphics[width=1.6cm]{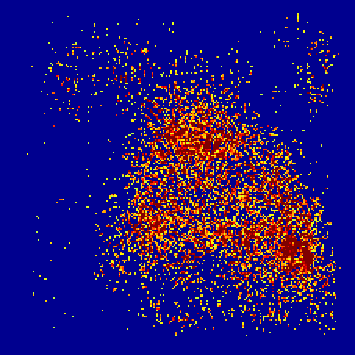}}
\end{minipage}
\hfill
\begin{minipage}{0.06\linewidth}
    \centerline{\includegraphics[width=1.6cm]{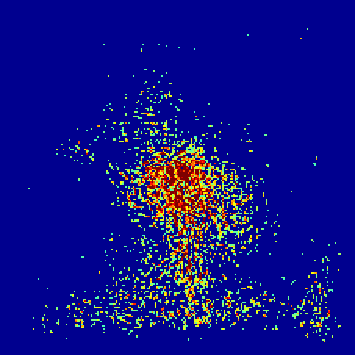}}
\end{minipage}
\vfill
\begin{minipage}{0.06\linewidth}
    \centerline{\scalebox{0.53}{(H) Raw saliency}}
    \centerline{\scalebox{0.53}{maps (CNN2+CNN3)}}
\end{minipage}
\hfill
\begin{minipage}{0.06\linewidth}
    \centerline{\includegraphics[width=1.6cm]{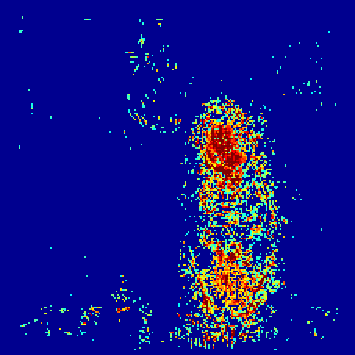}}
\end{minipage}
\hfill
\begin{minipage}{0.06\linewidth}
    \centerline{\includegraphics[width=1.6cm]{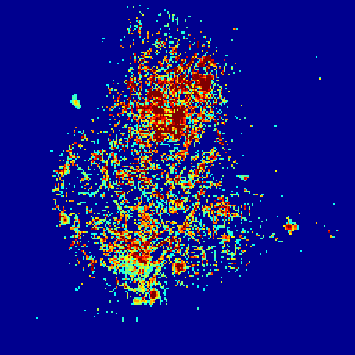}}
\end{minipage}
\hfill
\begin{minipage}{0.06\linewidth}
    \centerline{\includegraphics[width=1.6cm]{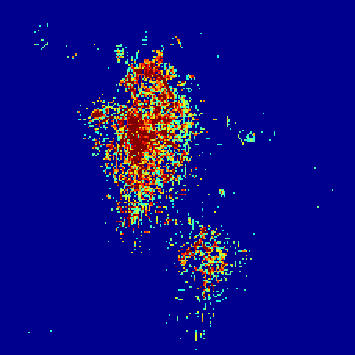}}
\end{minipage}
\hfill
\begin{minipage}{0.06\linewidth}
    \centerline{\includegraphics[width=1.6cm]{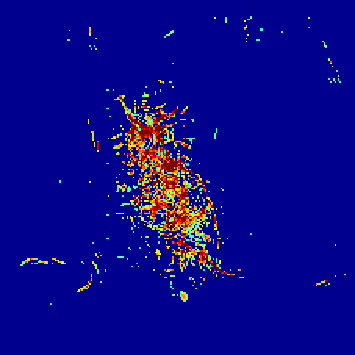}}
\end{minipage}
\hfill
\begin{minipage}{0.06\linewidth}
    \centerline{\includegraphics[width=1.6cm]{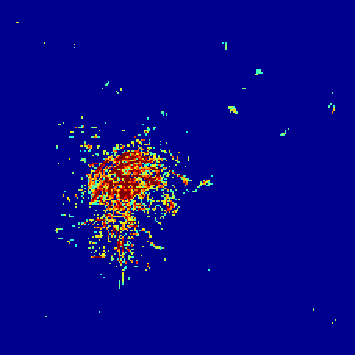}}
\end{minipage}
\hfill
\begin{minipage}{0.06\linewidth}
    \centerline{\includegraphics[width=1.6cm]{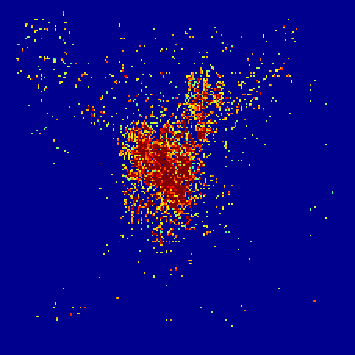}}
\end{minipage}
\hfill
\begin{minipage}{0.06\linewidth}
    \centerline{\includegraphics[width=1.6cm]{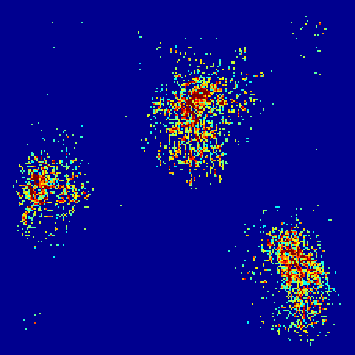}}
\end{minipage}
\hfill
\begin{minipage}{0.06\linewidth}
    \centerline{\includegraphics[width=1.6cm]{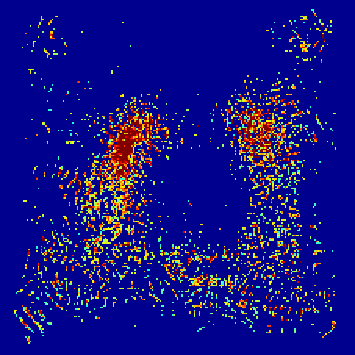}}
\end{minipage}
\hfill
\begin{minipage}{0.06\linewidth}
    \centerline{\includegraphics[width=1.6cm]{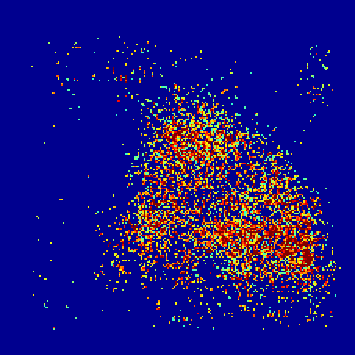}}
\end{minipage}
\hfill
\begin{minipage}{0.06\linewidth}
    \centerline{\includegraphics[width=1.6cm]{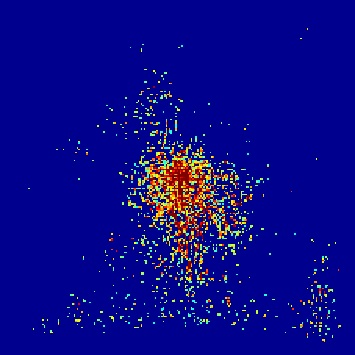}}
\end{minipage}
\vfill
\begin{minipage}{0.06\linewidth}
    \centerline{\scalebox{0.55}{(I) Smoothed}}
    \centerline{\scalebox{0.55}{saliency maps}}
\end{minipage}
\hfill
\begin{minipage}{0.06\linewidth}
    \centerline{\includegraphics[width=1.6cm]{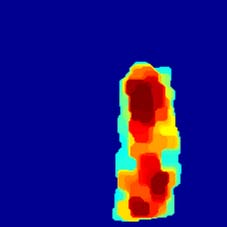}}
\end{minipage}
\hfill
\begin{minipage}{0.06\linewidth}
    \centerline{\includegraphics[width=1.6cm]{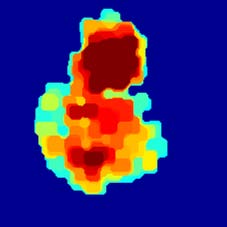}}
\end{minipage}
\hfill
\begin{minipage}{0.06\linewidth}
    \centerline{\includegraphics[width=1.6cm]{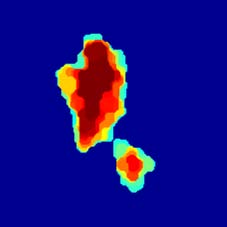}}
\end{minipage}
\hfill
\begin{minipage}{0.06\linewidth}
    \centerline{\includegraphics[width=1.6cm]{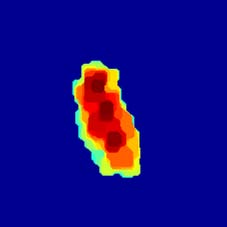}}
\end{minipage}
\hfill
\begin{minipage}{0.06\linewidth}
    \centerline{\includegraphics[width=1.6cm]{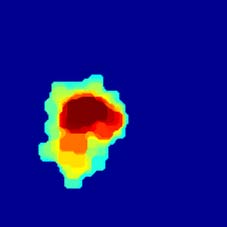}}
\end{minipage}
\hfill
\begin{minipage}{0.06\linewidth}
    \centerline{\includegraphics[width=1.6cm]{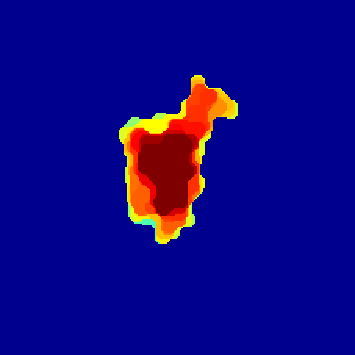}}
\end{minipage}
\hfill
\begin{minipage}{0.06\linewidth}
    \centerline{\includegraphics[width=1.6cm]{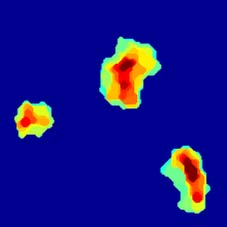}}
\end{minipage}
\hfill
\begin{minipage}{0.06\linewidth}
    \centerline{\includegraphics[width=1.6cm]{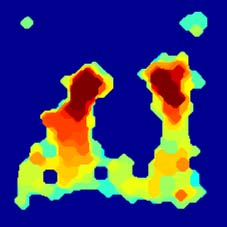}}
\end{minipage}
\hfill
\begin{minipage}{0.06\linewidth}
    \centerline{\includegraphics[width=1.6cm]{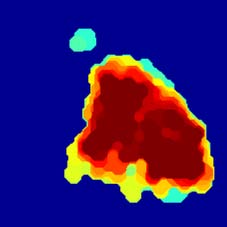}}
\end{minipage}
\hfill
\begin{minipage}{0.06\linewidth}
    \centerline{\includegraphics[width=1.6cm]{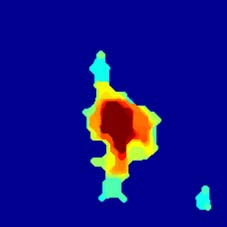}}
\end{minipage}
\vfill
\begin{minipage}{0.06\linewidth}
    \centerline{\scalebox{0.55}{(J) Refined}}
    \centerline{\scalebox{0.55}{Saliency maps}}
\end{minipage}
\hfill
\begin{minipage}{0.06\linewidth}
    \centerline{\includegraphics[width=1.6cm]{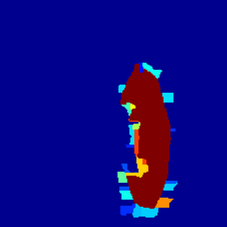}}
\end{minipage}
\hfill
\begin{minipage}{0.06\linewidth}
    \centerline{\includegraphics[width=1.6cm]{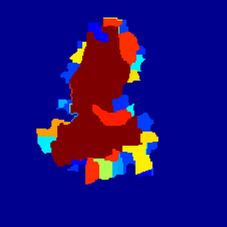}}
\end{minipage}
\hfill
\begin{minipage}{0.06\linewidth}
    \centerline{\includegraphics[width=1.6cm]{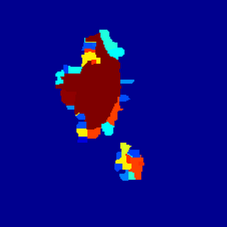}}
\end{minipage}
\hfill
\begin{minipage}{0.06\linewidth}
    \centerline{\includegraphics[width=1.6cm]{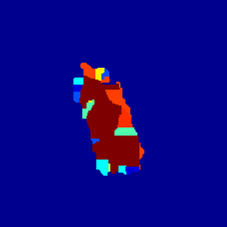}}
\end{minipage}
\hfill
\begin{minipage}{0.06\linewidth}
    \centerline{\includegraphics[width=1.6cm]{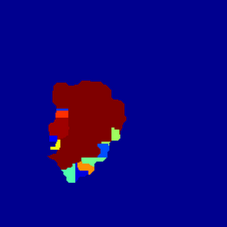}}
\end{minipage}
\hfill
\begin{minipage}{0.06\linewidth}
    \centerline{\includegraphics[width=1.6cm]{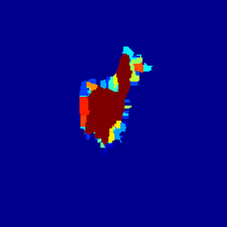}}
\end{minipage}
\hfill
\begin{minipage}{0.06\linewidth}
    \centerline{\includegraphics[width=1.6cm]{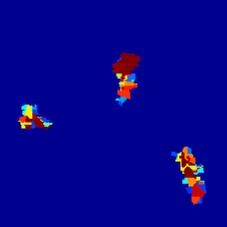}}
\end{minipage}
\hfill
\begin{minipage}{0.06\linewidth}
    \centerline{\includegraphics[width=1.6cm]{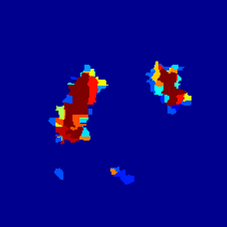}}
\end{minipage}
\hfill
\begin{minipage}{0.06\linewidth}
    \centerline{\includegraphics[width=1.6cm]{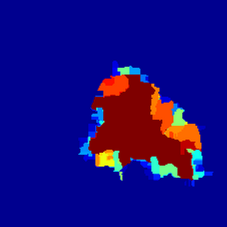}}
\end{minipage}
\hfill
\begin{minipage}{0.06\linewidth}
    \centerline{\includegraphics[width=1.6cm]{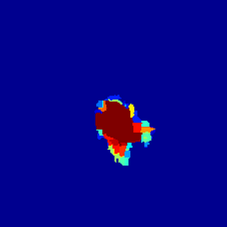}}
\end{minipage}
\vfill
\begin{minipage}{0.06\linewidth}
    \centerline{\scalebox{0.55}{(K) SaliencyCut \cite{cheng2011global}}}
\end{minipage}
\hfill
\begin{minipage}{0.06\linewidth}
    \centerline{\includegraphics[width=1.6cm]{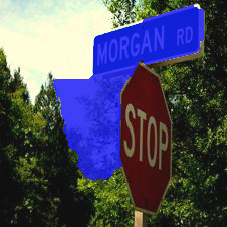}}
\end{minipage}
\hfill
\begin{minipage}{0.06\linewidth}
    \centerline{\includegraphics[width=1.6cm]{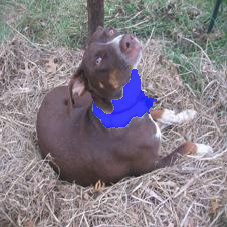}}
\end{minipage}
\hfill
\begin{minipage}{0.06\linewidth}
    \centerline{\includegraphics[width=1.6cm]{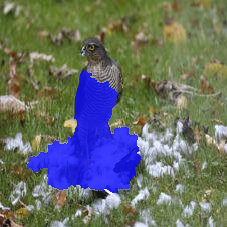}}
\end{minipage}
\hfill
\begin{minipage}{0.06\linewidth}
    \centerline{\includegraphics[width=1.6cm]{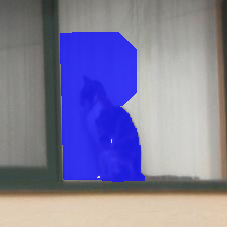}}
\end{minipage}
\hfill
\begin{minipage}{0.06\linewidth}
    \centerline{\includegraphics[width=1.6cm]{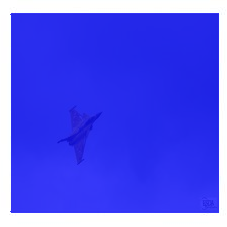}}
\end{minipage}
\hfill
\begin{minipage}{0.06\linewidth}
    \centerline{\includegraphics[width=1.6cm]{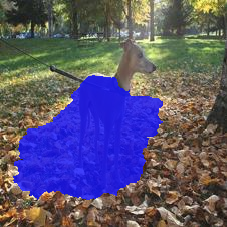}}
\end{minipage}
\hfill
\begin{minipage}{0.06\linewidth}
    \centerline{\includegraphics[width=1.6cm]{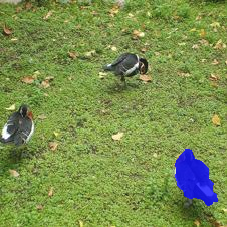}}
\end{minipage}
\hfill
\begin{minipage}{0.06\linewidth}
    \centerline{\includegraphics[width=1.6cm]{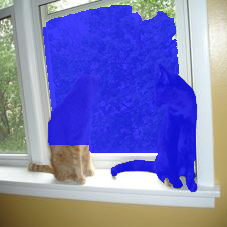}}
\end{minipage}
\hfill
\begin{minipage}{0.06\linewidth}
    \centerline{\includegraphics[width=1.6cm]{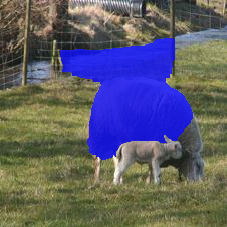}}
\end{minipage}
\hfill
\begin{minipage}{0.06\linewidth}
    \centerline{\includegraphics[width=1.6cm]{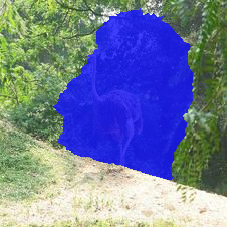}}
\end{minipage}
\vfill
\begin{minipage}{0.06\linewidth}
    \centerline{\scalebox{0.55}{(L) Our}}
    \centerline{\scalebox{0.55}{Segmentation}}
\end{minipage}
\hfill
\begin{minipage}{0.06\linewidth}
    \centerline{\includegraphics[width=1.6cm]{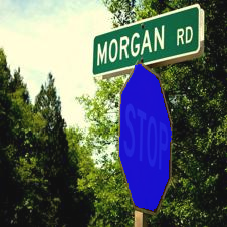}}
\end{minipage}
\hfill
\begin{minipage}{0.06\linewidth}
    \centerline{\includegraphics[width=1.6cm]{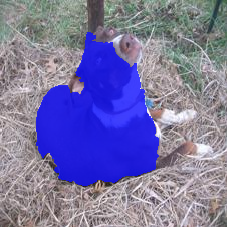}}
\end{minipage}
\hfill
\begin{minipage}{0.06\linewidth}
    \centerline{\includegraphics[width=1.6cm]{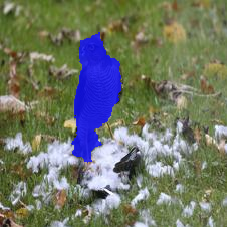}}
\end{minipage}
\hfill
\begin{minipage}{0.06\linewidth}
    \centerline{\includegraphics[width=1.6cm]{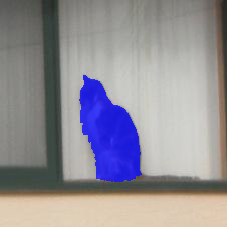}}
\end{minipage}
\hfill
\begin{minipage}{0.06\linewidth}
    \centerline{\includegraphics[width=1.6cm]{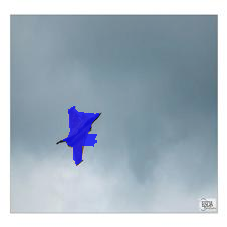}}
\end{minipage}
\hfill
\begin{minipage}{0.06\linewidth}
    \centerline{\includegraphics[width=1.6cm]{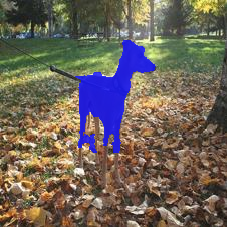}}
\end{minipage}
\hfill
\begin{minipage}{0.06\linewidth}
    \centerline{\includegraphics[width=1.6cm]{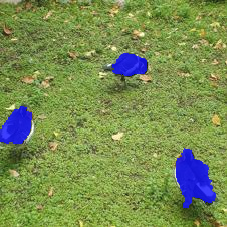}}
\end{minipage}
\hfill
\begin{minipage}{0.06\linewidth}
    \centerline{\includegraphics[width=1.6cm]{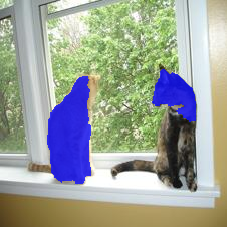}}
\end{minipage}
\hfill
\begin{minipage}{0.06\linewidth}
    \centerline{\includegraphics[width=1.6cm]{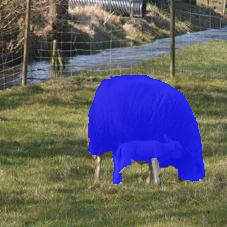}}
\end{minipage}
\hfill
\begin{minipage}{0.06\linewidth}
    \centerline{\includegraphics[width=1.6cm]{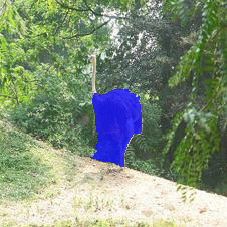}}
\end{minipage}
\vfill
\caption{Saliency Results of MS COCO (Column 1 to 5) and Pascal (Column 6 to 10). (A) original images, (B) masked images, (C) Region Contrast saliency maps \cite{cheng2011global} (D) DCNN based saliency maps by using \cite{Oxford-cnn-2014}, (E) to (H) raw saliency maps using CNN1, CNN2, CNN3 and CNN2 + CNN3, (I) smoothed saliency maps of (H) using image dilation and erosion, (J) refined saliency maps of (I), (K) segmentation using SaliencyCut \cite{cheng2011global} and (L) our segmentation results based on (J).}
\label{Fig:Saliency}
\end{figure*}

Similarly, we also use PR-curves and $F_{\beta}$ to evaluate the saliency and segmentation performance on Pascal VOC 2012 database. From Fig.~\ref{Fig:PRCurve_Pascal}, we can see that the proposed methods are significantly better than \cite{Oxford-cnn-2014}, and the DCNNs that make use of masked images yield comparable performance as \cite{cheng2011global}. As shown in Fig.~\ref{Fig:F_Pascal}, our methods still give slightly better $F_{\beta}$  values for both saliency detection and segmentation than \cite{cheng2011global} but the difference between them is not significant. This may be partially attributed to the poor DCNN models in the Pascal dataset, which is fine-tuned by only a very small number of in-domain images.
In Fig.~\ref{Fig:Saliency}, we also select several Pascal images to show the saliency and segmentation results (Column 6 to 10). Some of these examples have suggested that our methods are able to handle the images that contain multiple objects.

\section{Conclusion}
\label{sec_conc}
In this paper, we have proposed several novel DCNN-based methods for object saliency detection and image segmentation. The methods may utilize both original training images and masked images to train several DSCNNs. For each test image, we firstly recognize for the image class label, and then we can use any of the these DCNNs to generate a saliency map. Specifically, we attempt to reduce a cost function defined to measure the class-specific objectness of each image, and we back-propagate the corresponding error signals all way to the input layer and use the gradient of inputs to revise the input images. After several iterations, the difference between the original input images and the revised images is calculated as a saliency map. The saliency maps can be used to initialize an image segmentation algorithm to derive  the final segmentation results. We have evaluated our methods on two benchmark tasks, namely MS COCO \cite{lin2014microsoft} and Pascal VOC 2012 \cite{Everingham10}.
Experimental results  have shown that our proposed methods can generate high-quality salience maps, clearly outperforming many existing methods. In particular, our DCNN-based approaches excel on many difficult images, containing complex background, highly-variable salient objects, multiple objects, and very small objects.

\section*{Acknowledgement}

This work is partially supported by a research donation from iFLYTEK Co., Ltd., Hefei, China. We acknowledge NVIDIA for donating a Tesla    K40 GPU card under the Academic Partnership Program. The first author is supported by a scholarship from China Scholarship Council (CSC).


{\small
\bibliographystyle{ieee}
\bibliography{CNN_Saliency}
}

\end{document}